\documentclass{article}
\usepackage[final]{my_nips_2017}
\makeatletter
\renewcommand{\@notice}{}
\makeatother

\usepackage[utf8]{inputenc} 
\usepackage[T1]{fontenc}    
\usepackage{hyperref}       
\usepackage{url}            
\usepackage{booktabs}       
\usepackage{amsfonts}       
\usepackage{nicefrac}       
\usepackage{microtype}      
\usepackage{graphicx}

\usepackage{caption}
\usepackage{subcaption}

\title{Power-of-Two Quantization-Aware-Training (PoT-QAT) in Large Language Models (LLMs)}

\author{
  Mahmoud Elgenedy \\
  Department of Computer Science, Stanford University\\
  \texttt{melgened@stanford.edu} \\
}

\usepackage{enumitem}
\setlist{nosep} 

\renewenvironment{abstract}
{\vspace{-1em} 
 \begin{center}
   \bfseries Abstract
 \end{center}
 \vspace{-0.5em} 
 \small 
 \begin{quote}
}
{\end{quote}
 \vspace{-1em} 
}

\renewenvironment{abstract}
{\vspace{-1em}
 \begin{center}
   \bfseries Abstract
 \end{center}
 \vspace{-0.5em}
 \small
}
{\par\vspace{-1em}}

\begin{document}


\setlength{\textfloatsep}{8pt plus 1pt minus 2pt}
\setlength{\intextsep}{7pt plus 1pt minus 2pt}
\vspace{-2em} 

\maketitle
\vspace{-2em} 


\begin{abstract}
In Large Language Models (LLMs), the number of parameters has grown exponentially in the past few years, e.g., from 1.5 billion parameters in GPT-2 to 175 billion in GPT-3 to possibly more than trillion in higher versions. This raises a significant challenge for implementation, especially for Edge devices. Unlike cloud computing, memory and processing power for Edge devices are very limited, which necessitates developing novel ideas to make such applications feasible. In this work, we investigate compressing weights with a special quantization that limits numbers to only power-of-two (PoT). This helps save a huge amount of memory as only exponents need to be stored, more importantly, it significantly reduces processing power by replacing costly multiplication with low cost bit shifting. To overcome performance loss due to this strict quantization, we investigate Quantization Aware Training (QAT) to enhance performance through additional training. Results on GPT-2 124M show a major enhancement for quantized PoT model after additional training, with a perplexity enhancement of $66\%$ and BERT-Score loss to baseline GPT-2 of $1\%$. The memory saving is estimated to be $87.5\%$ while the inference speed is expected to be 3-10x faster with PoT quantization versus full-precision.

\end{abstract}

\section{Introduction}
\vspace{-.7em}
Quantization is a very effective method to optimize and reduce Deep Neural Networks DNN complexity, where numbers are represented with lower precision to help reduce both memory and processing requirements [1]. In this proposal, for the LLM models, we investigate restricting weights to power-of-two (PoT), aiming at a significant reduction in memory and computations. Moreover, to enhance the performance, we extend our work in [2] by exploring Quantization Aware Training (QAT). For other quantization approaches for LLM with QAT, refer to [3].

While quantization in neural networks is widely researched [4-6], only few studies investigated power-of-two quantization [7-10]. Specifically, restricting weights to PoT is investigated in a couple of studies for general DNNs [7-8] and with some focus on image applications using Convolutional Neural Networks (CNNs) [9]. However, for LLMs applications, very limited researches are available including a recent publication [10] that proposes PoT with post training quantization (PTQ) similar to the approach we proposed in our earlier study in [2].

\vspace{-.7em}
\section{Dataset and Model}
\begin{figure} 
    \centering
    \includegraphics[width=.55\textwidth]{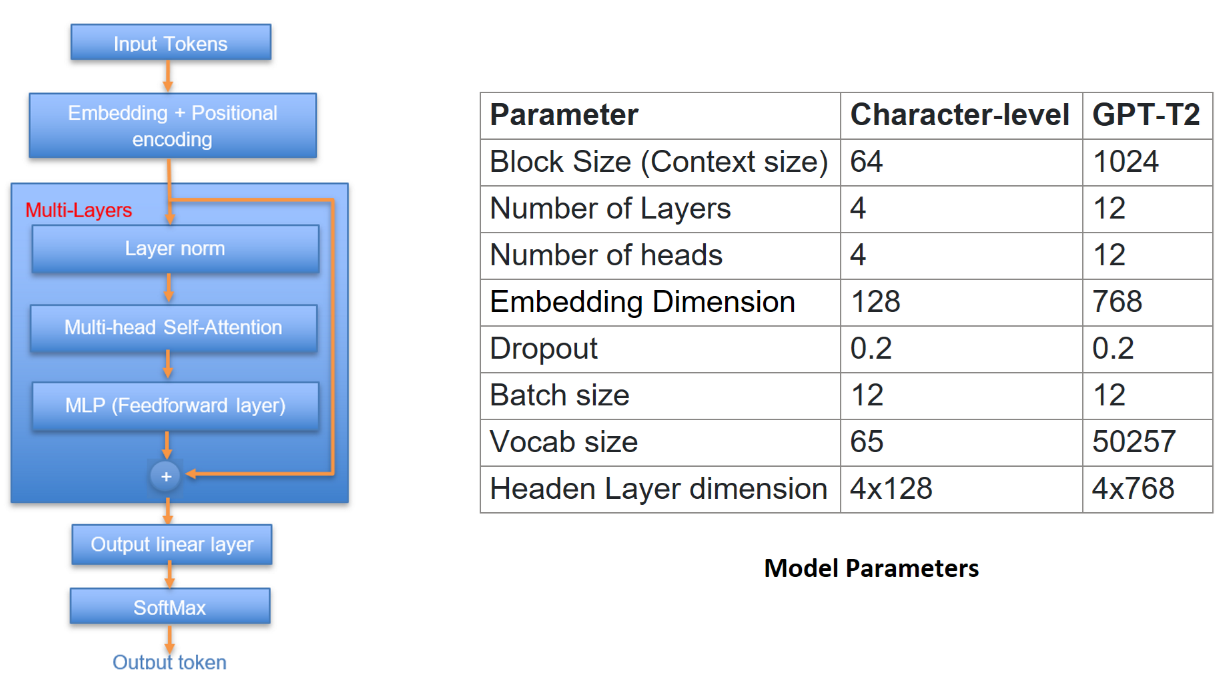}
    \caption{LLM Model illustration and parameters}
    \label{model-params}
\end{figure}

NanoGPT [11] is a simple and fast repository for training and fine tuning Generative Pre-trained Transformer GPT. The design follows the GPT-2 model and can reproduce GPT-2 on OpenWebText dataset. Moreover, it supports various settings, allowing fast trials and smaller datasets, including Shakespeare dataset. For preliminary results, we consider the Shakespeare dataset, partitioned as 80\%-10\%-10\% for training-val-testing, and focuses mainly on character-based model with model/parameters shown in figure \ref{model-params}. This is to help with short turn around testing and fine tuning, with more focus on the algorithm and correct implementation of the QAT for PoT. For more realistic results, we extended our PoT-QAT results to GPT-2 model. The baseline model is the 124M parameters pretrained model [12]. The QAT is fine tuned using same dataset used for GPT-2 training, OpenWebText dataset [13].

\vspace{-.7em}
\section{Quantization Aware Training}
\vspace{-.7em}
There are two different approaches for DNNs quantization: Post-Training Quantization (PTQ) and Quantization-Aware Training (QAT). In PTQ, training is done completely in floating point high precision, then weights and activations are quantized into lower precision during inference, no further tuning after quantization. PTQ is less complex but suffers higher performance loss.

On the other hand, in QAT, quantization is part of the training process, where fake quantize modules (that simulate the quantization loss by quantizing numbers followed by dequantizing/rescaling to floating range) are inserted between different operations, allowing training to fine tune the parameters considering quantization loss. Note that some of those operations like rounding are not differentiable which is a challenge to compute gradients for backward propagation. A common trick used is called straight-through estimator (STE) which treats the steps of rounding as identity only in the backward path, allowing gradients to flow through. STE approximation is proven to work well in practice and enables the model to converge and adapt to quantization noise during training [14]. 

QAT is computationally more expensive than PTQ but very effective to recover degradation due to quantization using additional training. Moreover, QAT is very critical when lower bit-widths like INT4 are used, or when models are sensitive to quantization noise, such as CNNs or transformers.
Since we are examining here a very aggressive quantization technique (PoT), we believe QAT would be a very effective approach to enhance the large performance degradation observed for PTQ [2].

\begin{figure} 
    \centering
    \includegraphics[width=.85\textwidth]{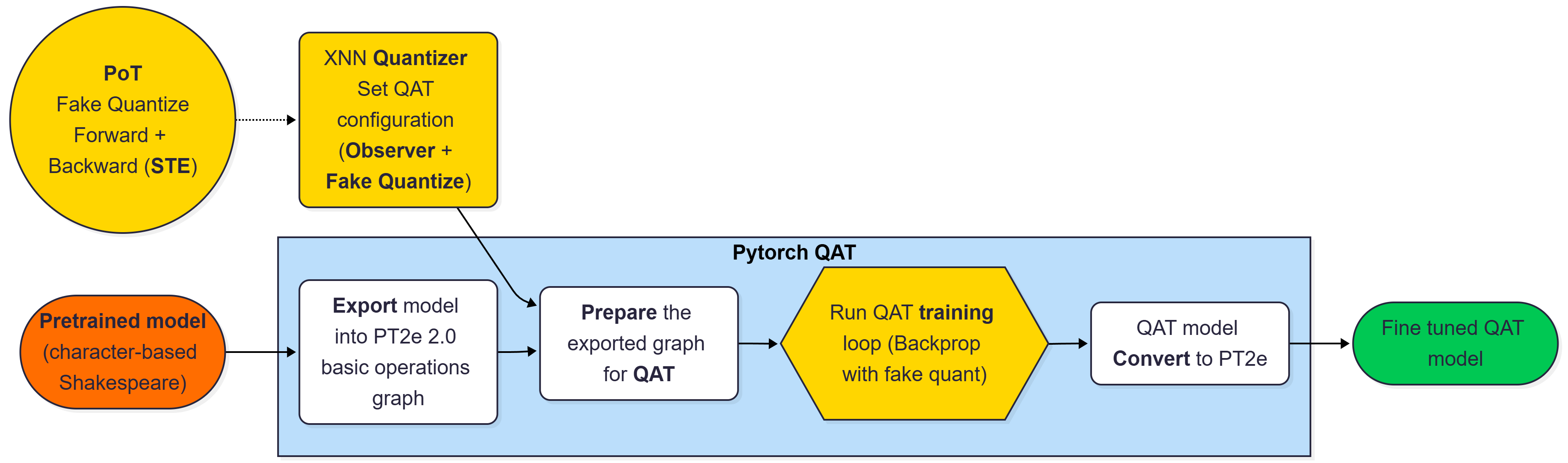}
    \caption{PyTorch QAT Flow}
    \label{pytorch_qat}    
\end{figure}

We use most recent PyTorch 2 Export Mode quantization framework [15-16] as it is more flexible and easily scalable. As in figure \ref{pytorch_qat}, the flow of QAT in PyTorch starts with exporting the model into a graph of basic operations. Next in prepare stage, fake quantizer modules are inserted between each two operations, based on configuration provided through the quantizer block. Inside quantizer we can configure fake quantize operation, quantization type and bit widths, in addition to signal statistics. In particular, the so called observer block is used to depict signal statistics (like min and max), then fake quantize can use these statistics to scale and quantize/dequantize signals. After prepare, the model is now ready for training to tune the weights with quantization effect. Lastly, after training is done, a convert process is used to remove observers and only keep fake quantization modules for inference.

\vspace{-.5em}
\vspace{-.5em}
\section{Power of Two Quantization}
\vspace{-.5em}
Power-of-two quantization is an aggressive quantization approach that limits quantized integer numbers to be only power of two. Multiplication (which is essential part of element-wise, matrix multiplication or convolution) is a major part of any DNN processing complexity. When one of the arguments has numbers that are restricted to be PoT, the multiplication operation turns into simple bit shifting which significantly reduces the computational complexity. Table \ref{quant-table} shows the formulas of the PoT versus Uniform affine quantization. The drawback of the PoT is the expected performance loss, which is the main motivation to introduce QAT trying to overcome that loss.

\begin{figure} 
    \centering
    \includegraphics[width=.75\textwidth]{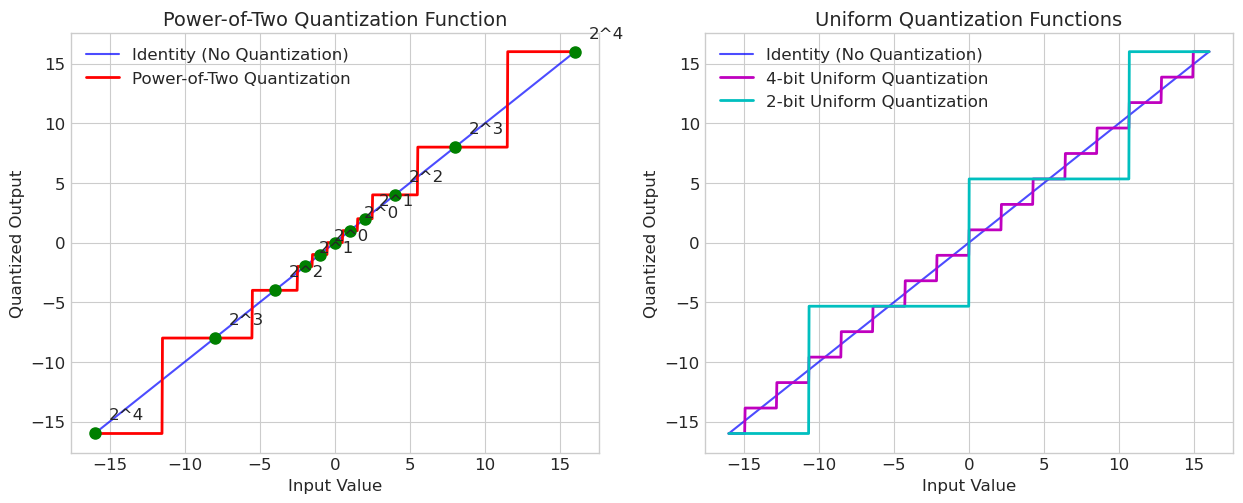}
    \caption{Power-of-Two (PoT) Versus Uniform Quantization}
    \label{pot_quant_STE}    
\end{figure}

As in uniform quantization, operation of PoT is not differentiable, and hence we also need to approximate the PoT into differentiable operation in the backward path. We examined the STE approach which seems to be really a good approximation and based on initial experiments, LLM model is able to converge. In figure \ref{pot_quant_STE}, we show the quantization output of both PoT and uniform quantization. PoT is a logarithmic approach where steps are not uniform (this can be beneficial for large outliers as a side benefit of PoT). STE assumes identity function in the backward path, i.e., output is same as input (the blue straight line in the figure).

\begin{table}
\caption{Types of Quantization}
\label{quant-table}

\begin{tabular}{ |p{3.0cm}||p{6.7cm}|p{3cm}| } 
 \hline
 Quantization type & Conversion equation (input x, and output y) & Use case\\
 \hline
 Asymmetric (Affine)
&   $y=round(x/scale) + zp$  & Activations\\
 Power of Two (PoT) &$ y = 2^{clip(round(\log_2(x/scale)))}$ & PoT Weights\\
 
 \hline
\end{tabular}
\end{table}
\vspace{-.5em}
\section{Experiments and Performance Evaluation}
\vspace{-.5em}

\subsection{Character based model with Shakespeare dataset}
For initial trials, we investigated the QAT performance of the character based LLM, where weights are quantized as uniform versus PoT (note that to benefit from PoT, only weights are needed to be in PoT). In table \ref{training-perf} we show training/validation loss convergence during training. We first trained the model on floating point for $15000$ iterations with $LR=10^{-3}$ which lowered cross entropy loss from little above $4$ to $~1.6$. We then load the trained model inside the QAT and train ($LR=.5*10^{-5}$) for both uniform and PoT quantization. The table shows results for uniform quantization 4-bits [-8, 7], versus 9-levels PoT. The results show clear convergence for both techniques, and proves that PoT can be enhanced with QAT. Specifically, the performance suffers an initial drop once applying quantization (from $~1.6$ baseline to above $1.8$ loss), then converges back after some training iterations.

\begin{table}
\caption{Nano-GPT Character Based LLM QAT Performance. Uniform quantization versus PoT}
\label{training-perf}

\begin{tabular}{ |p{3.0 cm}||p{2.15 cm}|p{2.15 cm}|p{2.15 cm}|p{2.15 cm}| } 
 \hline
 Iteration  & \multicolumn{2}{c|}{\bf{Uniform Quantization}} & \multicolumn{2}{c|}{\bf{Power-of-Two}} \\
 \cline{4-5}\cline{2-3}
  & Training & Validation & Training & Validation  \\
 \hline
 15500 &  1.61635  & \bf{1.84743} & 1.65916 & \bf{1.82383} \\
 19500 & 1.56602 & 1.68807 &  1.63777 & 1.75756 \\
 \bf{22500} &  1.61594 & \bf{1.59317} & 1.66232 &  \bf{1.68218} \\
 
 \hline
\end{tabular}
\end{table}
\vspace{-.5em}

\subsection{GPT-2 124M model with openwebtext data-set}

We extended results of the PoT-QAT on the GPT-2 124M model. We use different metrics to analyze performance, including train/val loss convergence curves, perplexity, and BERT-Score.

\subsubsection{Training/Validation loss convergence}

\begin{figure}
\centering

\begin{subfigure}[b]{0.5\textwidth}
  \includegraphics[width=1\linewidth]{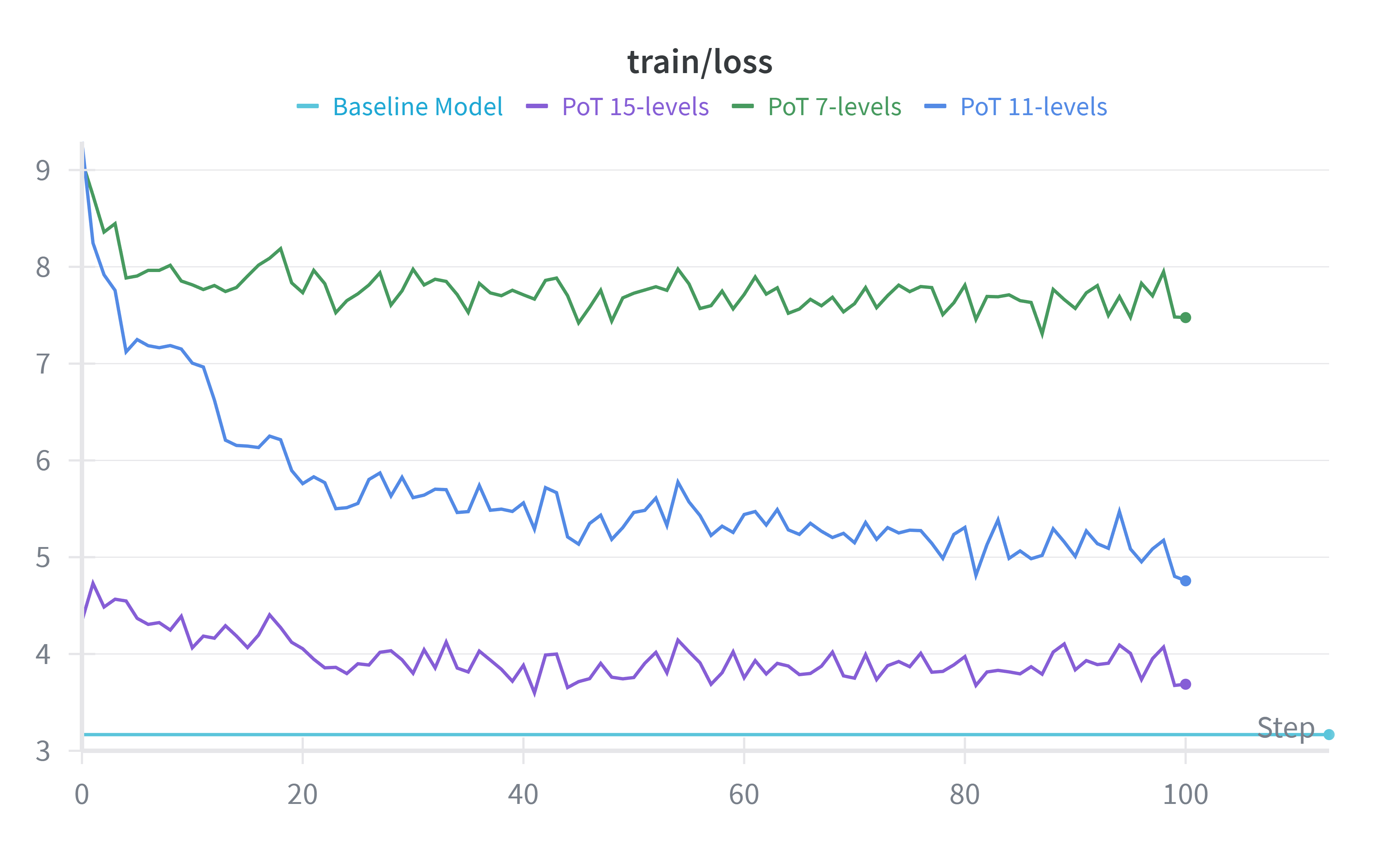}
  \caption{}
  \label{fig:conv1} 
\end{subfigure}

\medskip 
\begin{subfigure}[b]{0.5\textwidth}
  \includegraphics[width=1\linewidth]{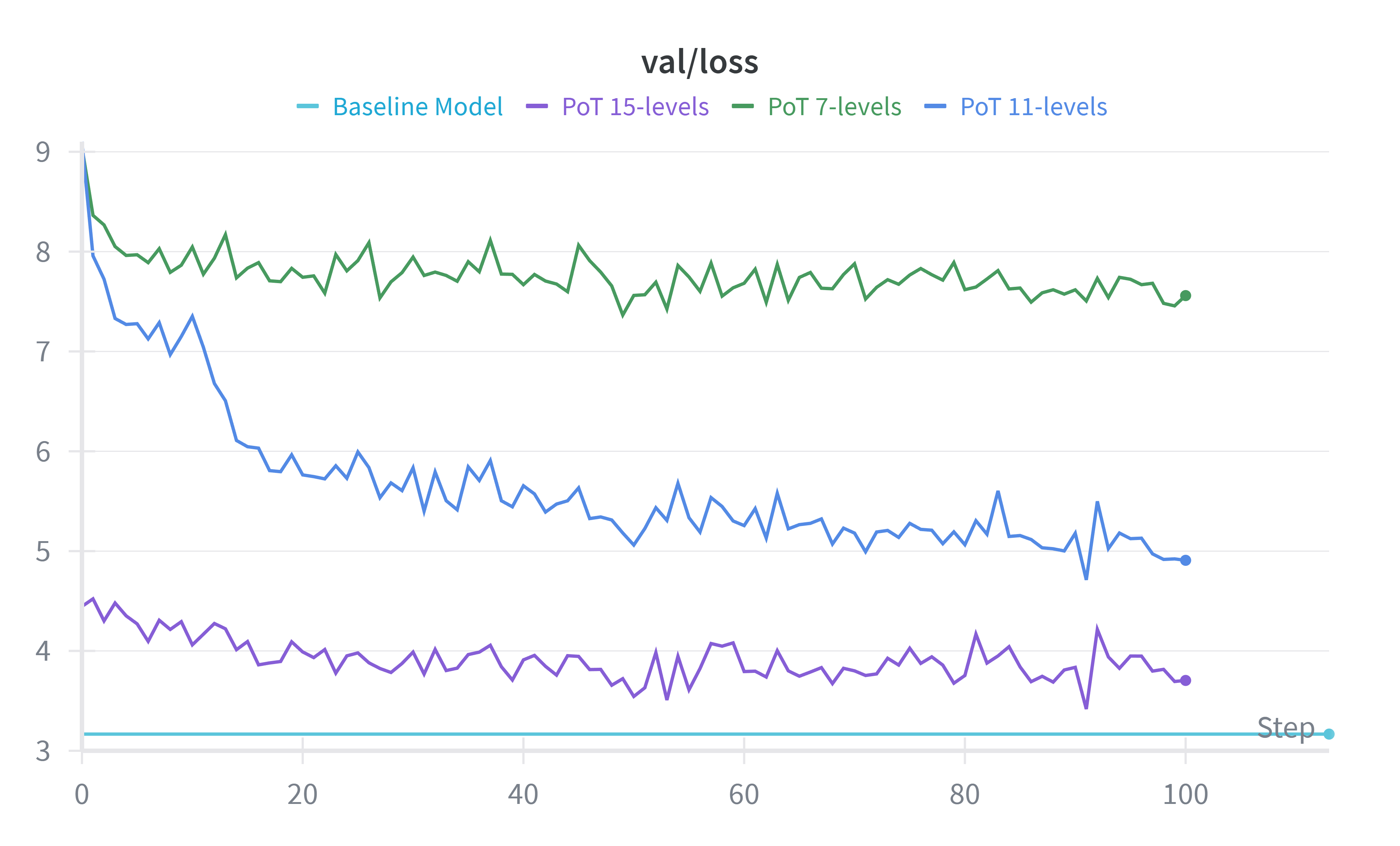}
  \caption{}
  \label{fig:conv2}
\end{subfigure}

\caption[PoT QAT convergence ]{%
Quantization Aware Training convergence for different PoT settings. Each step is 5 training step iterations.
(a) Training loss versus iterations. 
(b) Validation loss versus iterations.}

\end{figure}

As shown in figures \ref{fig:conv1} and \ref{fig:conv2}, the training and validation loss successfully converges. We tried different values of learning rates LRs and $.5e-5$ was the successful pick. The context length is 64, while the batch size is chosen to be 12 (we swept over different batch sizes 12, 24, 48, 64). We tried different resolution of PoT quantization (7, 11 and 15 levels). PoT with 15 levels ($[-2^7, 2^7]$) is able to close the performance gap degradation compared to baseline model. Compared to other PoT variants, PoT 15 levels shows initial small bias, while a slight overfitting is observed toward end of training. On the other hand, PoT with 7 and 11 levels show relatively large bias which is enhanced with training especially for the 11 levels (further training for 11 levels may gain more enhancement).




\vspace{-.5em}
\subsubsection{Perplexity}

Perplexity is a measure of how accurate the model can predict next token. In other words, it measures how the model is able to learn the data distribution. As the name implies, lower perplexity means model is able to predict with less ambiguity. It is calculated as the exponential of the cross entropy loss $l_{ce}$, i.e., perplexity is $\exp({l_{ce})} $

To measure cross-entropy, we average over multiple iterations of batches captured from val set. The cross-entropy $l_{ce}(t,y)$ with input $k-$dimension logits $t \in \mathbb{R}^k$ and target class $y$ is defined as,
 $l_{ce}(t,y) = -\log P(y;t) = -\log \left( \frac{\exp(t_y)}{\sum_{s} \exp(t_s)} \right) $

In Table \ref{perplexity-table}, we show the perplexity of quantized PoT for Post training PTQ, i.e., PoT quantization without additional training (prior results) versus the introduced quantization aware training QAT. The results show a major enhancement with training (QAT). For example, for the 15-levels PoT (equivalently 4 bits), the QAT is able to lower the perplexity from 90 (of PTQ) to 30, which is a significant enhancement (over $66\%$), and is very close to baseline model perplexity of 23.7.

\begin{table}
\caption{Perplexity performance of PoT post training quantization PTQ versus PoT quantization aware training quantization QAT}
\label{perplexity-table}

\begin{tabular}{ |p{2.5cm}|p{1.cm}||p{2cm}|p{2cm}| p{2cm}| p{2cm}|}
 \hline
 \multicolumn{2}{|c||}{quantization range} & \multicolumn{2}{c|}{PoT - PTQ} & \multicolumn{2}{c|}{\bf{PoT - QAT}} \\
 \hline
 Integer range & N bits & $l_{ce}$  &Perplexity & $l_{ce}$& Perplexity\\
 \hline
 Baseline float32 & 32 bits & 3.167&  23.73 & 3.167 & 23.167\\
 $[-2^{3},2^{3}]$ & 3 bits & 9.01 &  8184 & 7.42 & 1669\\
 $[-2^{5},2^5]$   & 4 bits    & 8.76 &   6374  & 4.71 & 111 \\
 $[-2^{7},2^7]$&   \bf{4 bits}  & 4.5   & \bf{90} & 3.417 & \bf{30.47}\\

 \hline
\end{tabular}
\end{table}

\subsubsection{BERT-Score}

\begin{figure} 
    \centering
    \includegraphics[width=.6\textwidth]{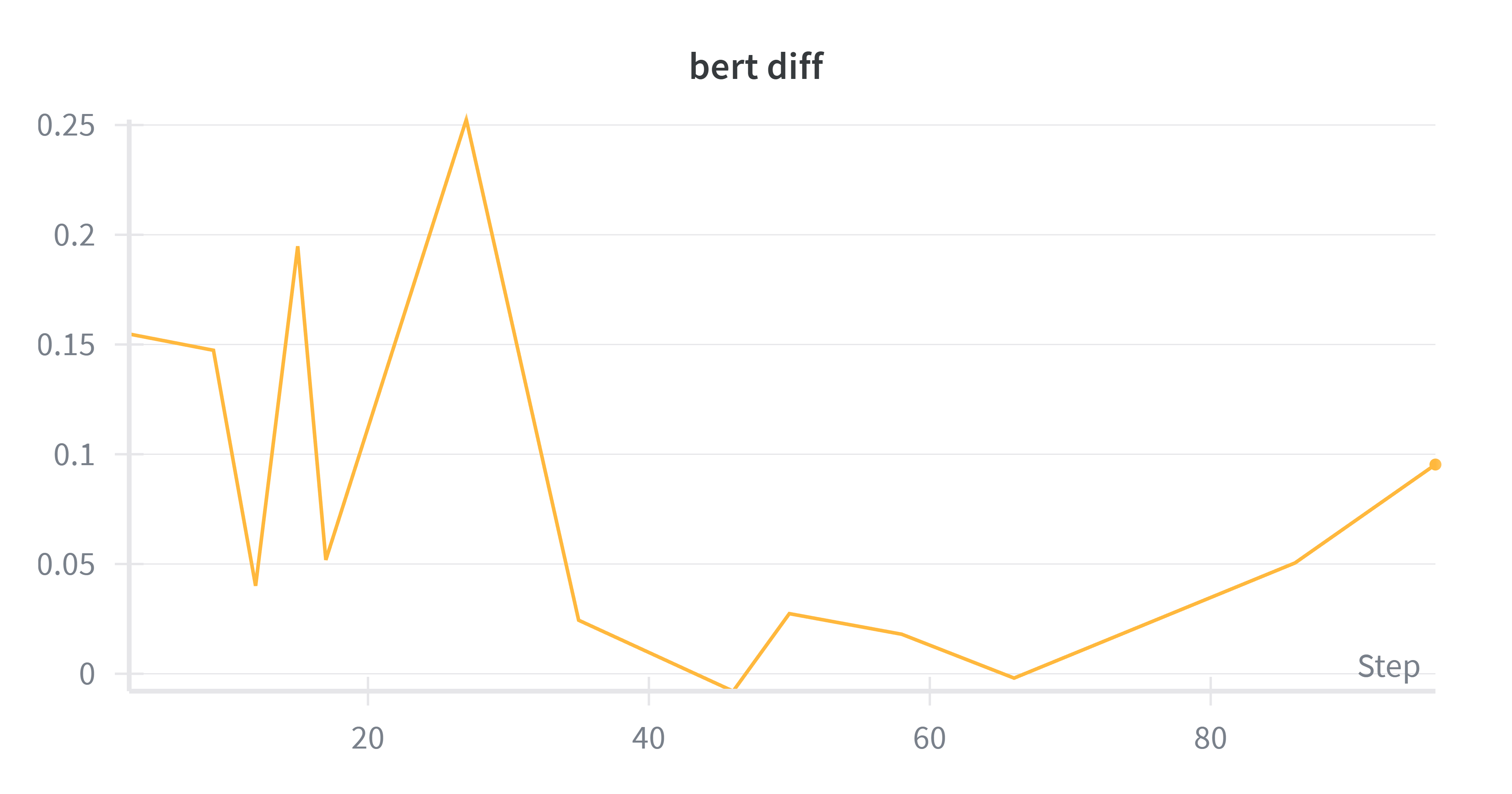}
    \caption{BERT-Score delta for PoT-QAT quantized vs baseline floating-point, at val loss $\approx3.78$}
    \label{bert}    
\end{figure}

BERT-Score is a similarity score for each token in the candidate sequence with each token in the reference sequence, where token similarity is done using contextual embeddings instead of exact matching [17]. BERT-Score is a good indication to human judgment. We use Hugging Face evaluate package to calculate the BERT-Score, with base model "roberta-base". We average over 10 iterations, where for each iteration we generate a new text and compute the BERT-Score. In figure \ref{bert}, we plot the BERT-Score difference between baseline floating point and quantized model with training iterations, only reported at checkpoints where better validation is observed. The BERT-Score is computed for baseline versus reference data, and similarly for the quantized model. The average BERT-Score of the baseline model is about 0.2, which is not very high, but expected considering GPT-2 124M. It is clear from the curve that the gap between quantized and baseline GPT-2 is becoming closer with training progressing, until reaching smallest BERT-Score difference ($0.0019$ difference $\approx 1 \%$ loss) around validation iteration 66 (training step 330). It worth noting that best BERT-Score is not at best validation error (best BERT-Score is aligned with validation error $\approx 3.78$ while best validation error is 3.41), which may indicate some overfitting toward the end of the training loop. This is very useful to pick the best model, where all metrics should be considered. For reference we show in figures \ref{fig:Ng1} and \ref{fig:Ng2} in the appendix a sample of generated text by both base and quantized models.

\vspace{-.5em}

\vspace{-.5em}

\section{Memory and Computational Savings}

Table \ref{tab:complexity_saving}, shows a summary of the memory and computation savings of using 4-bits PoT quantization. The memory saving is an outcome of using smaller bit-width (4-bits in PoT versus 32-bits of baseline float32), which is similar to any 4-bits quantization technique. However, the computation saving is a combination of saving from using smaller bit-width weights in all matrix multiplications (2-5x faster) in addition to the bit shifting operation that replaces multiplication (additional 1.5-2x faster).
\vspace{-.5em}
\begin{table}[htbp]
\centering
\caption{GPT-2 124M: FP32 vs. 4-bit Power-of-Two Quantization}
\begin{tabular}{lcc}
\toprule
\textbf{Metric} & \textbf{FP32} & \textbf{4-bit PoT} \\
\midrule
Model Size & 124 MB x 4 = 496 MB & 62 MB (87.5\% saving) \\
Inference Speed & 1x & 3-10x faster \\
\bottomrule
\end{tabular}
\label{tab:complexity_saving}
\end{table}

\vspace{-.5em}
\section{Conclusions and future work}
\vspace{-.5em}
We proposed power of two quantization for LLM weights which significantly reduces memory and computational complexity. We proposed using QAT so that additional training can help mitigate the performance degradation due to aggressive quantization. PoT functionality is implemented within PyTorch QAT framework for nano-GPT code. Results for GPT-2 124M show significant enhancement for the PoT with QAT compared to quantization without training (PTQ). Perplexity for 15 levels PoT gained $66\%$ enhancement over PTQ. The BERT-Score is also showing clear enhancement with training loop, where difference of BERT-Score versus baseline model approaches $1\%$. Future works includes experimenting the PoT-QAT for bigger models like Llama and Qwen. Also, it will be very interesting to try the algorithm on device and measure the actual complexity saving.

\section*{Contributions}
\begin{itemize}
    \item Implementing the QAT framework for nano-GPT LLM model, and configure/optimize model parameters (like LR) and quantization parameters to achieve convergence.
    \item Implement The PoT differentiable class that Includes both forward and backward functions.
    \item Integrating the PoT class inside the QAT framework (inheriting the fake quantize class).
    \item Achieving convergence for QAT for both uniform and PoT on the character level LLM model as first step.
    \item Extended our results for more realistic GPT-2 124M parameters. Achieved convergence for the PoT-QAT for different versions of PoT with different resolution.   
    \item Measured the performance using different metrics including train/val loss, perplexity, and BERT-Score.
    \item Finally, performance metrics show a great success for the QAT to help PoT to close the degradation gap compared to the baseline model ($66\%$ enhancement for perplexity over PTQ and $1\%$ loss of BERT-Score versus full precision)
\end{itemize}

\section*{References}

\medskip
\small

[1] What are Quantized LLMs?
"https://www.tensorops.ai/post/what-are-quantized-llms\#viewer-2tmd3"

[2] Elgenedy, Mahmoud. "Power-of-Two (PoT) Weights in Large Language Models (LLMs)." arXiv preprint arXiv:2506.00315 (2025).

[3] Liu, Zechun, Barlas Oguz, Changsheng Zhao, Ernie Chang, Pierre Stock, Yashar Mehdad, Yangyang Shi, Raghuraman Krishnamoorthi, and Vikas Chandra. "Llm-qat: Data-free quantization aware training for large language models." In Findings of the Association for Computational Linguistics: ACL 2024, pp. 467-484. 2024.

[4] Gholami, Amir, Sehoon Kim, Zhen Dong, Zhewei Yao, Michael W. Mahoney, and Kurt Keutzer. "A survey of quantization methods for efficient neural network inference." In Low-Power Computer Vision, pp. 291-326. Chapman and Hall/CRC, 2022.

[5] Krishnamoorthi, Raghuraman. "Quantizing deep convolutional networks for efficient inference: A whitepaper." arXiv preprint arXiv:1806.08342 (2018).

[6] Wu, Hao, Patrick Judd, Xiaojie Zhang, Mikhail Isaev, and Paulius Micikevicius. "Integer quantization for deep learning inference: Principles and empirical evaluation." arXiv preprint arXiv:2004.09602 (2020).

[7] Elhoushi, Mostafa, Zihao Chen, Farhan Shafiq, Ye Henry Tian, and Joey Yiwei Li. "Deepshift: Towards multiplication-less neural networks." In Proceedings of the IEEE/CVF conference on computer vision and pattern recognition, pp. 2359-2368. 2021.

[8] Przewlocka-Rus, Dominika, Syed Shakib Sarwar, H. Ekin Sumbul, Yuecheng Li, and Barbara De Salvo. "Power-of-two quantization for low bitwidth and hardware compliant neural networks." arXiv preprint arXiv:2203.05025 (2022).

[9] McDanel, Bradley, Sai Qian Zhang, H. T. Kung, and Xin Dong. "Full-stack optimization for accelerating cnns using powers-of-two weights with fpga validation." In Proceedings of the ACM International Conference on Supercomputing, pp. 449-460. 2019.

[10] Wang, Xinyu, Vahid Partovi Nia, Peng Lu, Jerry Huang, Xiao-Wen Chang, Boxing Chen, and Yufei Cui. "PoTPTQ: A Two-step Power-of-Two Post-training for LLMs." arXiv preprint arXiv:2507.11959 (2025).

[11] https://github.com/karpathy/nanoGPT, by Karpathy

[12] https://huggingface.co/openai-community/gpt2

[13] https://huggingface.co/datasets/Skylion007/openwebtext

[14] https://api.wandb.ai/links/byyoung3/eb1eylcu

[15] https://pytorch.org/docs/stable/quantization.html{\#}quantization-api-summary

[16] (prototype) PyTorch 2.0 Export Post Training Static Quantization — PyTorch Tutorials 2.0.1+cu117 documentation

[17] Zhang, Tianyi, Varsha Kishore, Felix Wu, Kilian Q. Weinberger, and Yoav Artzi. "Bertscore: Evaluating text generation with bert." arXiv preprint arXiv:1904.09675 (2019).

\section{Acknowledgments}
I have done all work myself, Mahmoud Elgenedy, and no other team members.
I would like to acknowledge my Instructors and TAs of CS230 for all their help and support.
The Latex template is slightly modified from, "CS230: Deep Learning, Winter, 2018, Stanford University, CA. (LateX template borrowed from NIPS 2017.)"
I got some help in basics questions from internet search and using LLMs.

\section*{Appendix}

\subsection{Codes}

All codes are uploaded to following github,

\href{https://github.com/melgenedy/Quantized_nano_GPT-QAT-POT.git}{\url{https://github.com/melgenedy/Quantized_nano_GPT-QAT-POT.git}}

\subsection{Some Snippets from Quantized model}

\begin{figure}[ht] 

    \label{quant_model}
    \centering
    \includegraphics[width=1.1\textwidth]{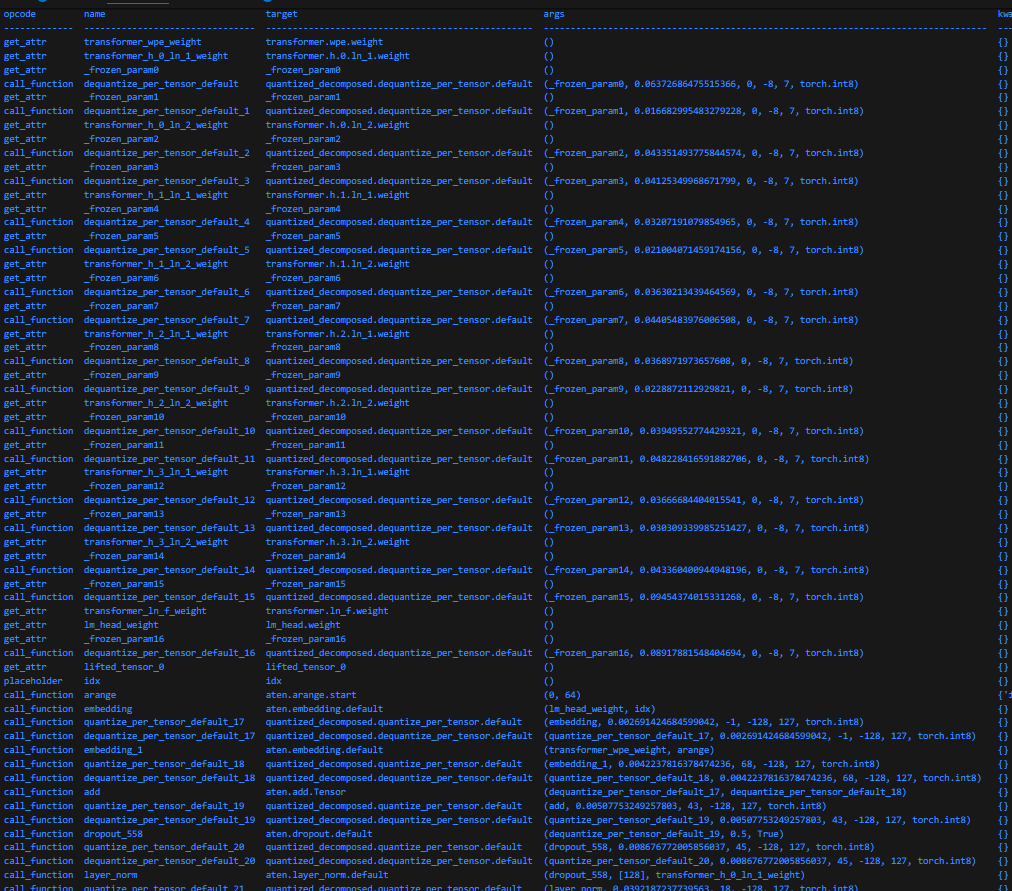}
    \caption{A snippet from the quantized model shown as print tabular}    
\end{figure}

\vspace{5.5em}

\subsection{Some Snippets from generated Quantized model versus baseline GPT-2 model}

\begin{figure}[ht]
\centering

\begin{subfigure}[b]{1\textwidth}
  \includegraphics[width=1.0\linewidth]{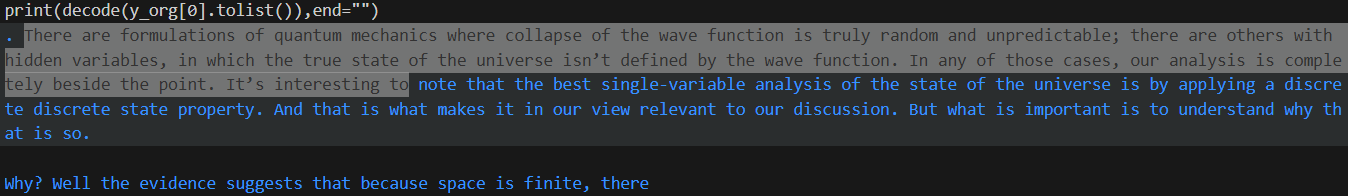}
  \caption{}
  \label{fig:Ng1} 
\end{subfigure}

\medskip 
\begin{subfigure}[b]{1.0\textwidth}
  \includegraphics[width=1\linewidth]{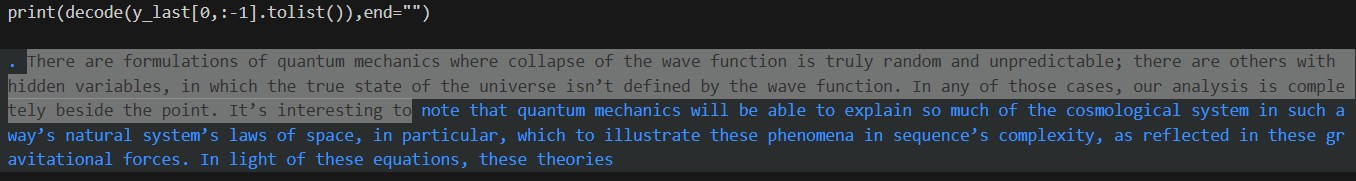}
  \caption{}
  \label{fig:Ng2}
\end{subfigure}

\caption[PoT QAT generated text ]{%
Generated text with context length 64 (highlighted), temperature = $0.8$. The generated text from quantized model at validation loss 3.76 seems very plausible compared to the baseline model. Note that baseline model is GPT-2 124M, which is relatively a small model compared to newer or larger parameters model, so the generated text even from base model is limited but is very useful to proof of concept.
(a) Baseline GPT-2 124M parameters, floating point. 
(b) PoT-QAT quantized model at val loss $\approx 3.7$.}

\end{figure}

\end{document}